# EKGNet: A 10.96μW Fully Analog Neural Network for Intra-Patient Arrhythmia Classification

Benyamin Haghi
*Electrical Engineering*
Caltech
Pasadena, CA, USA
ballahgh@caltech.edu

Lin Ma
*Electrical Engineering*
Caltech
Pasadena, CA, USA
lma5@caltech.edu

Sahin Lale
*Electrical Engineering*
Caltech
Pasadena, CA, USA
alale@caltech.edu

Anima Anandkumar
*Computer Science*
Caltech
Pasadena, CA, USA
anima@caltech.edu

Azita Emami
*Electrical Engineering*
Caltech
Pasadena, CA, USA
azita@caltech.edu

*Abstract*—We present an integrated approach by combining analog computing and deep learning for electrocardiogram (ECG) arrhythmia classification. We propose EKGNet, a hardware-efficient and fully analog arrhythmia classification architecture that achieves high accuracy with low power consumption. The proposed architecture leverages the energy efficiency of transistors operating in the subthreshold region, eliminating the need for analog-to-digital converters (ADC) and static random-access memory (SRAM). The system design includes a novel analog sequential Multiply-Accumulate (MAC) circuit that mitigates process, supply voltage, and temperature variations. Experimental evaluations on PhysioNet's MIT-BIH and PTB Diagnostics datasets demonstrate the effectiveness of the proposed method, achieving an average balanced accuracies of 95% and 94.25% for intra-patient arrhythmia classification and myocardial infarction (MI) classification, respectively. This innovative approach presents a promising avenue for developing low power arrhythmia classification systems with enhanced accuracy and transferability in biomedical applications.

## I. Introduction

The electrocardiogram (ECG) is crucial for monitoring heart health in medical practice [1], [2]. However, accurately detecting and categorizing different waveforms and morphologies in ECG signals is challenging, similar to other time-series data. Moreover, manual analysis is time-consuming and prone to errors. Given the prevalence and potential lethality of irregular heartbeats, achieving accurate and cost-effective diagnosis of arrhythmic heartbeats is crucial for effectively managing and preventing cardiovascular conditions [3], [4].

Deep neural network-based algorithms [5] are commonly used for ECG arrhythmia classification (AC) due to their high accuracy [6]. However, many of the current highly accurate arrhythmia classifiers that rely on neural networks (NN) require a large number of trainable parameters, often ranging from thousands to millions, to achieve their exceptional performance [6]–[11]. This poses a significant challenge when implementing these classifiers on hardware, as accommodating such a vast number of parameters becomes impractical. Consequently, existing algorithms are computationally intensive, particularly when compared to biological neural networks that operate with significantly lower energy requirements. As a result, designing low-power NN-AC systems poses significant computational challenges due to the computational demands involved.

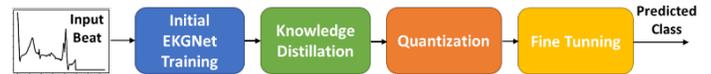

Fig. 1: Model training pipeline - EKGNet training and optimization.

Current approaches aim to tackle this either by (1) designing better AC algorithms, (2) better parallelism and scheduling on existing hardware such as graphics processing units (GPUs) or, (3) designing custom hardware. Previous studies [12]–[15] that concentrate on patient-specific arrhythmia classification on chip necessitate training neural networks individually for each patient, which significantly limits their potential applications. Moreover, most of the existing hardware development is with respect to digital circuits.

Analog computing in the subthreshold region offers potential energy efficiency improvements, eliminating the need for ADC and SRAM, in contrast to prior research that mainly focused on digital circuit implementations [16], [17]. This is particularly beneficial for ECG classification applications, which often face energy constraints in health monitoring devices [18]–[23]. Despite the challenges associated with analog circuits, such as susceptibility to noise and device variation, they can be effectively utilized for inferring neural network algorithms. The presence of inherent system noise in analog circuits can be leveraged to enhance robustness and improve classification accuracy, aligning with the desirable properties of AI algorithms [24]–[26].

In this paper, we propose EKGNet, a fully analog neural network with low power consumption (10.96μW) that achieves high balanced accuracies of 95% on the MIT-BIH dataset and 94.25% on the PTB dataset for intra-patient arrhythmia classification. To address the challenges of analog circuits, we design an integrated approach that combines AI algorithms and hardware design. By modeling the EKGNet as a Bayesian neural network using Bayes by Backprop [27], we incorporate analog noise and mismatches into the EKGNet model [28]. Knowledge distillation [29] is employed to further enhance the network's performance by transferring knowledge from ResNet18 [30] used as a teacher network to the EKGNet. We also propose an algorithm to conduct weight fine-tuning after quantization to improve hardware performance.

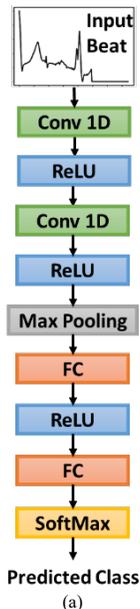
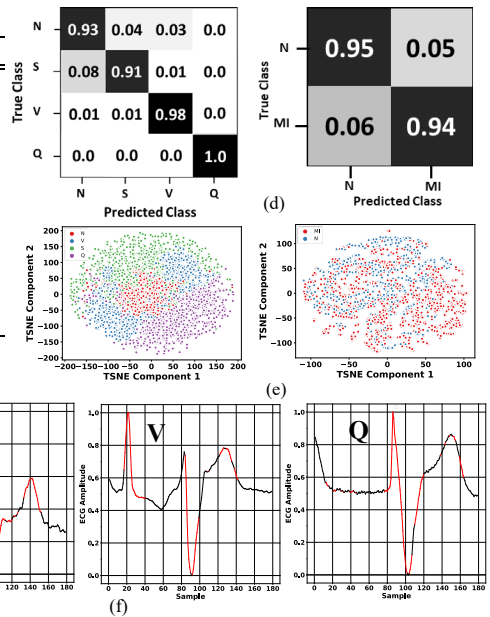

Fig. 2. (a) EKGNet architecture. (b) Table I: Mapping of beat annotations to AAMI EC57 categories. Table II: EKGNet details. (c) Algorithm 1: Fine-tuning of weights after quantization. (d) Confusion matrix for MIT-BIH (left) and PTB (right) classifications. (e) t-SNE visualization of learned representation for MIT-BIH (left) and PTB (right) classifications. Task labels are color-coded. (f) Colored sections highlight important segments in EKGNet predictions.

## II. DATASETS

In this work we utilize two databases; the PhysioNet MIT-BIH Arrhythmia dataset and PTB Diagnostic ECG dataset [31]–[33], for labeled ECG records. Specifically, we focused on ECG lead II. The MIT-BIH dataset included ECG recordings from 47 subjects, sampled at 360Hz, with beat annotations by cardiologists. Following the AAMI EC57 standard [34], beats were categorized into four categories based on annotations (Table I). The PTB Diagnostics dataset contained ECG records from 290 subjects, including 148 with myocardial infarction (MI), 52 healthy controls, and other subjects with different diseases. Each record in this dataset consisted of ECG signals from 12 leads, sampled at 1000Hz. Our analysis concentrated on ECG lead II and the MI and healthy control categories.

## III. METHODS

### A. Data Preparation

We extract beats from ECG recordings for classification by employing a straightforward and effective method [8]. Our approach avoids signal filtering or processing techniques that rely on specific signal characteristics. The extracted beats are of uniform length, ensuring compatibility with subsequent processing stages. The process involves resampling the ECG data to 125Hz, dividing it into 10-second windows, and normalizing the amplitude values between zero and one. We identify local maxima through zero-crossings of the first derivative and determine ECG R-peak candidates using a threshold of 0.9 applied to the normalized local maxima. The median of the R-R time intervals within the window provides the nominal heartbeat period (T). Each R-peak is associated with a signal segment of 1.2T length, padded with zeros to achieve a fixed length. The inputs are adjusted to fit our hardware input range of 0.6 V to 0.7 V (600 mV to 700 mV).

To address dataset imbalance, we divided the data into training and testing sets. For balanced representation, we excluded a specific number of beats for test: 3200 beats (800 beats per class) for the MIT-BIH and 2911 beats (809 healthy beats and 2102 MI beats) for the PTB dataset. The remaining beats underwent random oversampling [35], resulting in an augmented training dataset with an equal number of beats in each class. We ensured complete separation of training and testing data before augmentation to prevent overfitting. After augmentation, the training dataset consisted of 352,276 beats for the MIT-BIH (88,069 beats per class) and 16,800 beats for the PTB dataset (8,400 beats per class).

### B. EKGNet training

To implement the fully analog NN-AC, we optimized the software using a co-design approach. The hardware behavior was emulated in software by extracting a mathematical model of the Multiply-Accumulate (MAC) unit from circuit simulations. EKGNet, a convolutional neural network (CNN), was trained for ECG classification using the constructed ECG training set. During training, Bayes by Backprop [27] was utilized to model the standard deviation of weights ($w$) as derived hardware input-referred thermal noise ($\sigma = 0.0021090w^2 + 0.0002000w + 0.002355$)[1]. Hardware leakage noise ($\sim \mathcal{N}(0.0005\,V, 0.0001\,V)$) was integrated into the network's output. The training pipeline is depicted in Fig. 1, and the high-level architecture of EKGNet is shown in Fig. 2a and Table II. EKGNet consists of two 1-D convolutional layers, two ReLU activations, a max pooling layer, two fully connected layers, and a softmax layer [5]. For optimization, we employed Adam with $L_2$ regularization weight decay to optimize the cross-entropy loss [37]. Learning rate of α = 0.003 was used, which was halved every fifty epochs using a linear scheduler. This approach ensured that the trained weights remained within a small range suitable for implementation and improved linearity due to hardware noise characteristics (Fig. 5).

---

[1] The weights and coefficients are expressed in Volts.

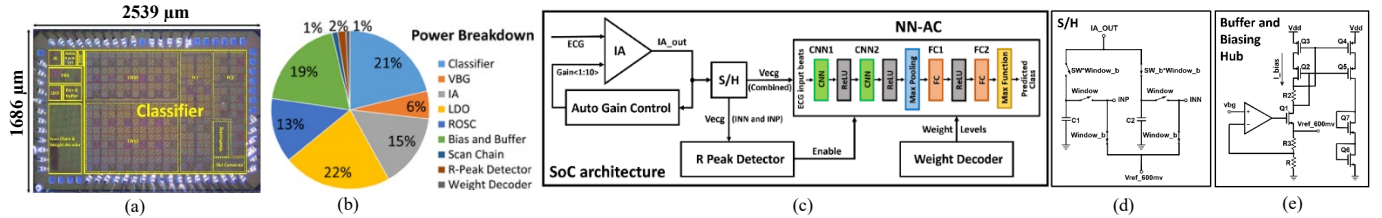

Fig. 3. (a) Die micrograph, (b) Power breakdown, (c) NN-AC and SoC Architectures, (d) Sample and Hold (S/H), (e) Buffer and biasing hub.

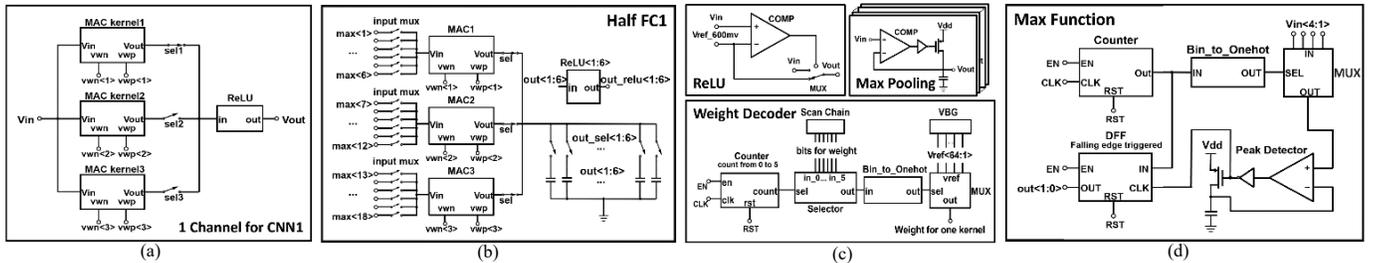

Fig. 4. (a) 1 channel for CNN1, (b) Half FC1, (c) ReLU, max pooling (18 parallel peak detectors), and weight decoder, (d) Max function.

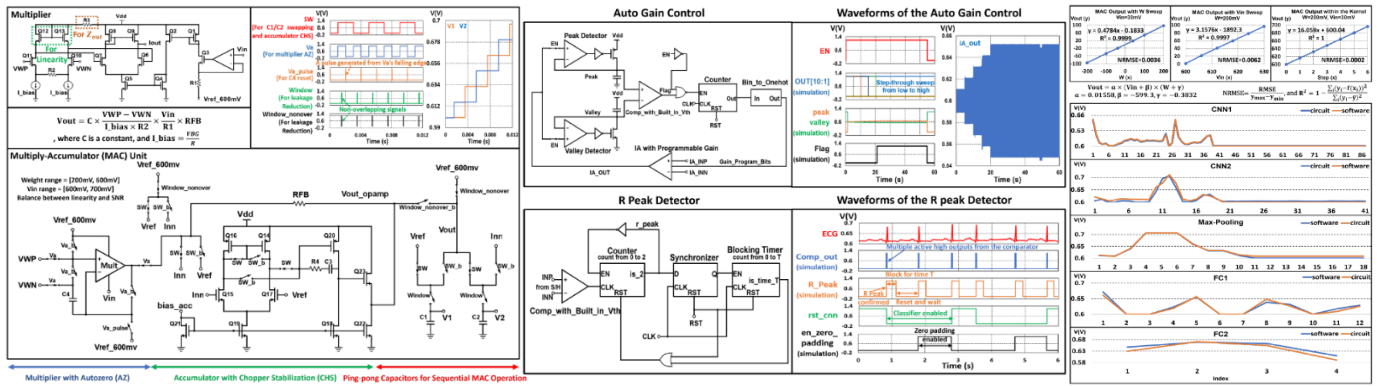

Fig. 5. (Left) Schematic, waveforms, and equation for the MAC unit, showing its tracking technique to reduce sensitivity to process parameter and temperature variations. (Middle) IA with automatic gain control, R-peak detector, and their measured waveforms. (Right) MAC characterization simulation results.

By applying knowledge distillation [29] to further train EKGNet, we observed a performance improvement of 1.5% on MIT-BIH dataset (resulting in 95% test accuracy) and 1.25% on PTB dataset (resulting in 94.25% test accuracy). Knowledge distillation involves transferring knowledge from a larger teacher network (ResNet18) with high test accuracies (99.88% for MIT-BIH and 100% for PTB datasets) to the smaller student network (EKGNet). Through experimentation, we determined that a temperature parameter value of 1.5 yielded optimal results, considering EKGNet's significantly fewer trainable parameters (336) compared to ResNet18 (~11 million).

To balance power consumption and accuracy, we used a 6-bit uniform quantization for the weights. Employing a fine-tuning technique, we iteratively adjusted a single weight by shifting it up or down one quantization level and evaluating its impact on performance (Fig. 2c, Algorithm 1). With this approach, we achieved the hardware performance of 94.88% and 94.10% on the MIT-BIH and PTB datasets, respectively.

### C. Model Interpretability

Interpreting machine learning algorithms, especially deep learning, in medical applications is a significant challenge [38]. We utilized t-SNE to visualize the learned representation by mapping high-dimensional vectors of the classified beats to a 2D space [39]. In Fig. 2e, we demonstrate clear separability between different classes using MIT-BIH and PTB datasets. Notably, only predicted class labels were used for colorization in the visualizations. To identify regions of input data that receive more attention from EKGNet during prediction, we selected a representative input beat from each category of the MIT-BIH dataset (Fig. 2f). Color-coded visual representations were employed to highlight segments of higher importance in EKGNet's predictions. By calculating the average Shapley value [40] across the entire beat, we selectively colored samples surpassing the threshold. Fig. 2f illustrates the most typical attribution pattern for ECG classification, aligning with established ECG abnormalities such as ST-segment elevation (STE) and pathological Q waves. However, some model attributions are less conclusive, and the highlighted areas may not perfectly align with clinical significance.

## IV. Hardware Architecture

The proposed hardware architecture includes a fully analog NN-AC and System-on-Chip (SoC) implementation (Fig. 3). The analog NN-AC, optimized for analog computing, has 336 parameters. Digitally assisted analog circuits are used for ReLU, max pooling, and max functions in the NN-AC. The SoC integrates power-on-reset, bandgap voltage reference, biasing hub, oscillator, scan chain, and low dropout regulators (LDO). An LDO with minimal output variations enhances the analog NN-AC's robustness against supply fluctuations. All circuits operate in the subthreshold region with strict duty cycle control for reduced power consumption.

TABLE III. COMPARISON OF SOFTWARE-ONLY ALGORITHMS

| MIT-BIH Dataset | Method | Conv. Layers | FC Layers | Parameters | Accuracy (%) |
|---|---|---|---|---|---|
| **This Paper (EKGNet)** | **Shallow CNN** | **2** | **2** | **336** | **95.00** |
| Acharya et al. [7] | Deep CNN | 3 | 3 | 19,805 | 94.03 |
| Kachuee et al. [8] | Deep Residual CNN | 11 | 2 | 98,757 | 93.40 |
| Yan et al. [11] | Deep CNN | 5 | 3 | 196,526 | 92.00 |
| Almahfuz et al. [9] | Deep CNN | 13 | 4 | 4,391,685 | 99.90 |
| This Paper (Teacher Net) | ResNet18 | 17 | 1 | 11M | 99.88 |
| **PTB Dataset** | | | | | |
| **This paper (EKGNet)** | **Shallow CNN** | **2** | **2** | **312** | **94.25** |
| Acharya et al. [7] | Deep CNN | 3 | 3 | 19,805 | 93.50 |
| Kachuee et al. [8] | Deep Residual CNN | 11 | 2 | 98,757 | 95.90 |
| Kojuri et al. [10] | Deep CNN, Resnet | 18, 40 | 0, 1 | 145,209; 5,001,842 | 95.60 |
| This Paper (Teacher Net) | ResNet18 | 17 | 1 | 11M | 100.00 |

TABLE IV. COMPARISON OF HARDWARE DESIGNS

| | This Paper | JSSC2014 [12] | TBCAS2020 [13] | TCASII2021 [14] | ISSCC2021 [15] |
|---|---|---|---|---|---|
| Process | **65 nm** | 90 nm | 0.18 μm | 0.18 μm | 65 nm |
| Area | **4.28** | 4.99 | 0.93 | 0.75 | 1.74 |
| Complete SoC | **Yes** | Yes | No | No | No |
| Computing Scheme | **Analog** | Digital | Digital | Digital | Digital |
| Require ADC | **No** | Yes | Yes | Yes | Yes |
| System VDD (V) | **1.55** | 0.5-1 (0.7-1 for SRAM) | N/A | N/A | N/A |
| Classifier VDD (V) | **1.2** | 0.5-1 | 1.8 | 1.8 | 0.75 |
| Test Dataset | **MIT-BIT & PTB** | In-house & MIT-BIH | MIT-BIH | MIT-BIH | MIT-BIH |
| Class Number | **4** | 2 | 4 | 5 | 2/5 |
| Intra-Patient | **Yes** | Yes | No, patient specific | No, patient specific | No, patient specific |
| Method | **CNN+FC** | MLC/SVM | NN (FC) | NN (FC) | CNN+FC |
| Accuracy | **94.88%(Arrythmia) 94.10%(MI)** | 95.8%(Arrythmia) 99%(MI) | 99.32% | 98% | 99.30% (2 class) 99.16% (5 class) |
| Accuracy On | **Test Data** | Train* & Test | Train* & Test | Train* & Test | Train* & Test |
| System Power (μW) | **67.07** | 102.2 | N/A | N/A | N/A |
| Classifier Power (μW) | **10.96** | 32.8 | 13.34 | 1.3 | 46.8 @1MHz 86.7 @2.5MHz |
| Leakage Power (μW) | **N/A** | N/A | N/A | Not Reported | 14.3 |

* The reported accuracy was higher than anticipated due to the incomplete exclusion of the training data.

To achieve overlapping CNN operations in hardware, three parallel MAC units are used with a 2-input-sample delay. CNN1 has six channels with ReLU activation. CNN2 employs charge redistribution for average pooling across all six channels, followed by ReLU activation. The first half of the fully connected layer (FC1) in Fig. 4b consists of 18 input signals undergoing MAC operations in three MAC units. The outputs are combined and sequentially output as six signals. FC2 follows the same design. The max function selects the node with the highest voltage from FC2, producing a 2-bit digital code representing the input ECG's arrhythmia class. The weight decoder synchronizes with NN-AC's control signals to convert digital codes to analog voltage levels. The fully analog NN-AC incorporates inputs from the sample and hold (S/H), enable signals from the R-peak detector, and weight levels from the weight decoder, generating the 2-bit digital output indicating the ECG's arrhythmia class.

Fig. 5 depicts the analog MAC unit. It consists of a multiplier and a current ($I_{out}$) proportional to their product. To reduce noise and cancel offsets, the multiplier incorporates autozero functionality. Linearity enhancement is achieved through the integration of an inverse hyperbolic tangent circuit. Resistor R3 is included to optimize the multiplier's output impedance, ensuring shift-invariance of the MAC. The accumulator converts $I_{out}$ into a voltage and stores it in the ping-pong capacitors. During each conversion, one capacitor acts as $V_{ref}$, while the other capacitor stores the updated voltage $V_{ref} + I_{out} \times RFB$. This sequential MAC operation scheme reduces hardware and power requirements compared to parallel operations. The accumulator utilizes chopper stabilization to mitigate offsets and noise, employing switches controlled by narrow window pulses to minimize the leakage effect. The equation in Fig. 5 shows that the MAC output depends solely on the weight, $V_{in}$, and device matching.

We propose an analog R-peak detector in the analog domain for beat extraction, specifically identifying the maximum peak of the ECG R wave. Using ECG gradients, the signal is sampled at a rate of 125 samples per second (S/s) with a sample and hold (S/H) circuit employing two ping-pong capacitors to preserve consecutive samples (Fig. 3d). In contrast to previous studies relying on digital R-peak detection, we introduce a digitally assisted analog R-peak detector (Fig. 5, middle). By exploiting the higher gradient of the R wave in the ECG waveform, we accurately locate R-peaks by comparing the gradient obtained from the S/H with a predefined threshold. To address noise issues, a Schmitt trigger is integrated into the comparator, utilizing two consecutive active high outputs to confirm the presence of an R-peak. Maintaining a constant input amplitude to the NN-AC is essential for achieving an optimal balance between the linearity of the signal. We propose an automatic gain control mechanism (Fig. 5, middle) to address challenges in the MAC unit and signal-to-noise ratio (SNR). The mechanism includes peak and valley detectors that measure the output amplitude of the instrumental amplifier (IA). A comparator compares the IA output with a target value using a predefined threshold. The IA gain is adjusted systematically from low to high until the comparator changes state, indicating the desired amplitude is achieved. To optimize performance, bias terms are eliminated, and the IA with automatic gain control ensures a consistent output amplitude.

V. EXPERIMENTAL RESULTS

The proposed design underwent simulation and fabrication using a 65nm process. Extensive optimization and characterization of MAC linearity were performed through simulations (Fig. 5, right). The achieved normalized root mean square errors (NRMSE) for the weights and $V_{in}$ were 0.0036 and 0.0062, respectively. Simulations also confirmed linearity within the kernel, resulting in an NRMSE of 0.0002. This ensures the MAC unit's linearity and shift-invariance, enabling linear operations in the CNN and FC layers. The mathematical model of the MAC, presented in Fig. 5, along with simulated intermediate signals within the NN-AC, demonstrate waveform similarity to the software implementation with minor errors.

Our NN-AC achieved a measured accuracy of 94.88% and 94.10% on the MIT-BIH and PTB intra-patient classifications, respectively. The power consumption of the proposed NN-AC is 10.96μW at a supply voltage of 1.2V. The overall SoC consumes 67.07μW at a supply voltage of 1.55V. Power consumption breakdown for the SoC is provided in Fig. 3b. Additionally, Tables III and IV summarize the performance of our system, demonstrating lower parameters and power consumption compared to previous software and hardware designs while maintaining comparable accuracy utilizing the intra-patient paradigm.

VI. CONCLUSION

We have developed a fully analog CNN-based architecture for accurate arrhythmia classification, using the MIT-BIH and PTB datasets. Our system achieves high accuracy and reduces power consumption by utilizing analog computing, eliminating the requirement for ADC and SRAM. The integration of a novel analog sequential MAC circuit effectively handles PVT variations. Experimental outcomes validate the efficacy of our architecture, offering a low-power solution for accurate arrhythmia classification in wearable ECG sensors.

ACKNOWLEDGMENT

This work was partially supported by the Carver Mead New Adventure Fund and Heritage Medical Research Institute at Caltech. We would like to express our gratitude to Wei Foo for his help in software and hardware validations. Additionally, we extend our thanks to James Chen and Katie Chiu for their contribution to hardware validation. Finally, we acknowledge Dr. Jialin Song and Dr. Yisong Yue for their collaboration on this project.